\documentclass[11pt]{article}
\usepackage{times,amsmath,amssymb,indentfirst,epsfig,longtable}

\newcommand{\dsoff}{\renewcommand{\baselinestretch}{1}\large\normalsize}

\begin{document}

\begin{titlepage}
\vspace*{1.7in}
\begin{center}
\hspace*{1.2cm}{\large{\bf Sub--Structural Niching in}}\\
\hspace*{1.2cm}{\large{\bf Non--Stationary Environments}}\\
\hspace*{1.2cm}~\\
\hspace*{1.2cm}~\\
\hspace*{1.2cm}{\bf Kumara Sastry}\\
\hspace*{1.2cm}{\bf Hussein A. Abbass}\\
\hspace*{1.2cm}{\bf David E. Goldberg}\\
\hspace*{1.2cm}~\\
\hspace*{1.2cm}~\\
\hspace*{1.2cm}~\\
\hspace*{1.2cm}IlliGAL Report No. 2004035\\
\hspace*{1.2cm}May, 2004\\
\vspace*{6cm} {\large
\hspace*{1.2cm}Illinois Genetic Algorithms Laboratory (IlliGAL)\\
\hspace*{1.2cm}Department of General Engineering\\
\hspace*{1.2cm}University of Illinois at Urbana-Champaign \\
\hspace*{1.2cm}117 Transportation Building \\
\hspace*{1.2cm}104 S. Mathews Avenue, Urbana, IL 61801 \\}
\end{center}
\end{titlepage}
\dsoff

\title{Sub--Structural Niching in Non--Stationary Environments}

\author{$^\dag$Kumara Sastry, $^\ddag$Hussein A. Abbass, and $^\dag$David
Goldberg\\
$^\dag$Illinois Genetic Algorithms Laboratory\\
University of Illinois at Urbana-Champaign\\
104 S.Mathews Ave., Urbana, IL 61801\\ 
\{kumara,deg\}@illigal.ge.uiuc.edu\\
$^\ddag$Artificial Life and Adaptive Robotics Laboratory\\
School of Information Technology and Electrical Engineering\\
University of New South Wales, Australian Defence Force Academy, Canberra, ACT 2600, Australia\\
h.abbass@adfa.edu.au}

\date{}
\maketitle

\begin{abstract}

Niching enables a {\em genetic algorithm} (GA) to maintain diversity in a population. It is
particularly useful when the problem has multiple optima where the aim is to find all or as many as
possible of these optima. When the fitness landscape of a problem changes overtime, the problem is
called non--stationary, dynamic or time--variant problem. In these problems, niching can maintain
useful solutions to respond quickly, reliably and accurately to a change in the environment. In
this paper, we present a niching method that works on the problem substructures rather than the
whole solution, therefore it has less space complexity than previously known niching mechanisms. We
show that the method is responding accurately when environmental changes occur.
\end{abstract}

\section{Introduction}

The systematic design of genetic operators and parameters is a challenging task in the literature.
Goldberg \cite{Gold02} used Holland's \cite{Holl75} notion of building blocks to propose a
design--decomposition theory for designing effective {\em genetic algorithms} (GAs). This theory is
based on the correct identification of substructures in a problem to ensure scalability and
efficient problem solving. The theory establishes the principles for effective supply, exchange and
manipulation of sub--structures to ensure that a GA will solve problems quickly, reliably, and
accurately. These types of GAs are called {\em competent\/} GAs to emphasize their robust behavior
for many problems.

A wide range of literature exists for competent GAs. This literature encompasses three broad
categories based on the mechanism used to unfold the substructures in a problem. The first category
is Perturbation techniques which work by effective permutation of the genes in such a way that
those belonging to the same substructure are closer to each other. Methods fall in this category
include the messy GAs \cite{Gold89a}, fast messy GAs \cite{Gold93}, gene expression messy GAs
\cite{Karg96}, linkage identification by nonlinearity check GA, linkage identification by
monotonicity detection GA \cite{Mun99}, dependency structure matrix driven GA \cite{Yu03}, and
linkage identification by limited probing \cite{Heck03}.

The second category is linkage adaptation techniques, where promoters are used to enable genes to
move across the chromosome; therefore facilitating the emergence of genes' linkages as in
\cite{Chen04}. The third category is probabilistic model building techniques, where a probabilistic
model is used to approximate the dependency between genes. Models in this category include
population-based incremental learning \cite{Balu94}, the bivariate marginal distribution algorithm
\cite{Peli99}, the extended compact GA (ecGA) \cite{Hari99}, iterated distribution estimation
algorithm \cite{Bosm99}, and the Bayesian optimization algorithm (BOA) \cite{Peli00}.

When the fitness landscape of a problem changes overtime, the problem is called non--stationary,
dynamic or time--variant problem. To date, there have been three main evolutionary approaches to
solve optimization problems in changing environments. These approaches are: (1) diversity control
either by increasing diversity when a change occurs as in the hyper--mutation method \cite{Cobb},
the variable local search technique \cite{Vava97} and others \cite{Bier99,Lin97}; or maintaining
high diversity as in redundancy \cite{Gold87,Coll97,Dag95}, random immigrants \cite{Gref92}, aging
\cite{Ghos98}, and the thermodynamical GAs \cite{Mori96}; (2) memory-based approaches by using
either implicit \cite{Gold87} or explicit \cite{Mori97} memory; and (3) speciation and
multi--populations as in the self-organizing-scouts method \cite{Bran01}.

Niching is a diversity mechanism that is capable of maintaining multiple optima simultaneously. The
early study of Goldberg, Deb and Horn \cite{Gold92b} demonstrated the use of niching for massive
multimodality and deception. Mahfoud \cite{Mahf95} conducted a detailed study of niching in
stationary (static) environments. Despite that many of the studies found niching particularly
useful for maintaining all the global optima of a problem, when the number of global optima grows,
the number of niches can grow exponentially.

In this paper, we propose a niching mechanism that is based on the automatic identification and
maintaining sub--structures in non--stationary problems. We incorporate bounded changes to both the
problem structure and the fitness landscape. It should be noted that if the environment changes
either unboundedly or randomly, on average no method will outperform restarting the solver from
scratch every time a change occurs. We use a dynamic version of the {\em extended compact genetic
algorithm} (ecGA) \cite{Hari99}, called the {\em dynamic compact genetic algorithm} (dcGA)
\cite{Abb04TR}. We show that the proposed method can respond quickly, reliably, and accurately to
changes in the environment. The structure of the paper is as follows: in the next section, we will
present dcGA and niching, then a feasibility study to test niching is undertaken followed by
experiments and discussions.

\section{Dynamic Compact Genetic Algorithm (dcGA)}

Harik \cite{Hari99} proposed a conjecture that linkage learning is equivalent to a good model that
learns the structure underlying a set of genotypes. He focused on probabilistic models to learn
linkage and proposed the ecGA method using the {\it minimum description length} ($\mathrm{MDL}$)
principle \cite{Riss78} to compress good genotypes into partitions of the shortest possible
representations. The $\mathrm{MDL}$ measure is a tradeoff between the information contents of a
population, called compressed population complexity (CPC), and the size of the model, called model
complexity (MC).

The CPC measure is based on Shannon's entropy \cite{shannon48}, $E(\chi_I)$, of the population
where each partition of variables $\chi_I$ is a random variable with probability $p_{i}$. The
measure is given by
\begin{equation}
E(\chi_I)=-\ C \sum_{i}^{\sigma} p_{i} \log p_{i}
\end{equation}
where $C$ is a constant related to the base chosen to express the logarithm and $\sigma$ is the
number of all possible bit sequences for the variables belonging to partition $\chi_I$; that is, if
the cardinality of $\chi_I$ is ${\nu_I}$, $\sigma = 2 ^{\nu_I}$. This measures the amount of
disorder associated within a population under a decomposition scheme. The $\mathrm{MDL}$ measure is
the sum of CPC and MC as follows
\begin{equation}
\mathrm{MDL} \ \ = N \sum_I \left( -\ C \sum_{i}^{\sigma} p_{i} \log p_{i} \right) + \log(N)
2^{\nu_I}
\end{equation}
With the first term measures CPC while the second term measures MC.

In this paper, we assume that we have a mechanism to detect the change in the environment.
Detecting a change in the environment can be done in several ways including: (1) re--evaluating a
number of previous solutions; and (2) monitoring statistical measures such as the average fitness
of the population \cite{Bran01}. The focus of this paper is not, however, on how to detect a change
in the environment; therefore, we assume that we can simply detect it. The {\em dynamic compact
genetic algorithm} (dcGA) works as follows:

\begin{enumerate}
\item Initialize the population at random with $n$ individuals;

\item If a change in the environment is being detected, do:

\begin{enumerate}
\item Re--initialize the population at random with $n$ individuals;

\item Evaluate all individuals in the population;

\item Use tournament selection with replacement to select $n$ individuals;

\item Use the last found partition to shuffle the building blocks (building block--wise crossover)
to generate a new population of $n$ individuals;

\end{enumerate}

\item Evaluate all individuals in the population;

\item Use tournament selection with replacement to select $n$ individuals;

\item Use the $\mathrm{MDL}$ measure to recursively partition the variables until the measure
increases;

\item Use the partition to shuffle the building blocks (building block--wise crossover) to generate
a new population of $n$ individuals;

\item If the termination condition is not satisfied, go to 2; otherwise stop.

\end{enumerate}

Once a change is detected, a new population is generated at random then the last learnt model is
used to bias the re--start mechanism using selection and crossover. The method then continues with
the new population. In ecGA, the model is re-built from scratch in every generation. This has the
advantage of recovering from possible problems that may exist from the use of a hill--climber in
learning the model.

In the original ecGA and dcGA, the probabilities are estimated using the frequencies of the bits
after selection. The motivation is that, after selection, the population contains only those
solutions which are good enough to survive the selection process. Therefore, approximating the
model on the selected individuals inherently utilizes fitness information. However, if explicit
fitness information is used, problems may arise from the magnitude of these fitness values or the
scaling method.

Traditional niching algorithms work on the level of the individual. For example, re-scaling the
fitness of individuals based on some similarity measures. These types of niching require the niche
radius, which defines the threshold beyond which individuals are dissimilar. The results of a
niching method are normally sensitive to the niche radius. A smaller niche radius would increase
the number of niches in the problem and is more suitable when multiple optima are located closer to
each other, while a larger niche radius would reduce the number of niches but will miss out some
optima if the optima are close to each other. Overall, finding a reasonable value for the niche
radius is a challenging task.

When looking at ecGA, for example, the variables in the model are decomposed into subsets with each
subset represents variables that are tight together. In a problem with $m$ building blocks and $n$
global optima within each building block, the number of global optima for the problem is $n^m$.
This is an exponentially large number and it will require an exponentially large number of niches.
However, since the problem is decomposable, one can maintain the niches within each building block
separately. Therefore, we will need only $n \times m$ niches. Obviously, we do not know in advance
if the problem is decomposable or not; that is the power of ecGA and similar models. If the problem
is decomposable, it will find the decomposition, we can identify the niches on the level of the
sub--structures, and we save unnecessary niches. If the problem is not decomposable, the model will
return a single partition, the niches will be identified on the overall solution, and we are back
to the normal case. ecGA learns the decomposition in an adaptive manner, therefore, the niches will
also be learnt adaptively.

We propose two types of niches in dcGA for dynamic environments. We will call them Schem1 and
Schem2 respectively. For each sub-structure (partition), the average fitness of each schema $s$ is
calculated for partition $\chi_I$ as follows:
\begin{equation}
\text{Fit}(s \in \chi_I) = \frac{\sum_{i}^{p} \grave{f}_{is}}{\sigma}
\end{equation}
\begin{equation}
\grave{f}_{is} = \left\{ \begin{array}{lr} \grave{fit}_i & if \ \ i \mapsto s \\ 0 & otherwise \\
\end{array} \right.
\end{equation}
where, $fit_i$ is the fitness value of individual $i$, $\grave{f}_{is}$ is the fitness value of
individual $i$ if the schema $s$ is part of $i$ and 0 otherwise. The schema fitness is calculated
in Schem1 using the previous equation. In Schem2, the schema fitness is set to zero if its original
value is less than the average population fitness. In theory, it is good to maintain all schemas in
the population. In practice, however, maintaining all schemas will disturb the convergence of the
probabilistic model. In addition, due to selection, some below average schemas will disappear
overtime. Therefore, the choice will largely depend on the problem in hand. The dcGA algorithm is
modified into Schem1 and Schem2 by calculating the probabilities for sampling each schema based on
the schema fitness rather than the frequencies alone. This re--scaling maintains schemas when their
frequencies is small but their average fitness is high.

\section{Feasibility of the method}

Before we proceed with the experiments in the non--stationary environment, we need to check the
performance of the method on a stationary function. The function we use here is trap--4. Trap
functions were introduced by Ackley \cite{Ack87} and subsequently analyzed in details by others
\cite{Deb93,Gold02,Thie93}. A trap function is defined as follows
\begin{equation}
trap_k = \left\{ \begin{array}{ll} high & \text{if } u=k \\ low - u * \frac{low}{k-1} & otherwise
\end{array} \right.
\end{equation}
where, $low$ and $high$ are scalars, $u$ is the number of 1's in the string, and $k$ is the order
of the trap function.

\begin{figure} [h!]
\begin{center}
 \epsfig{figure=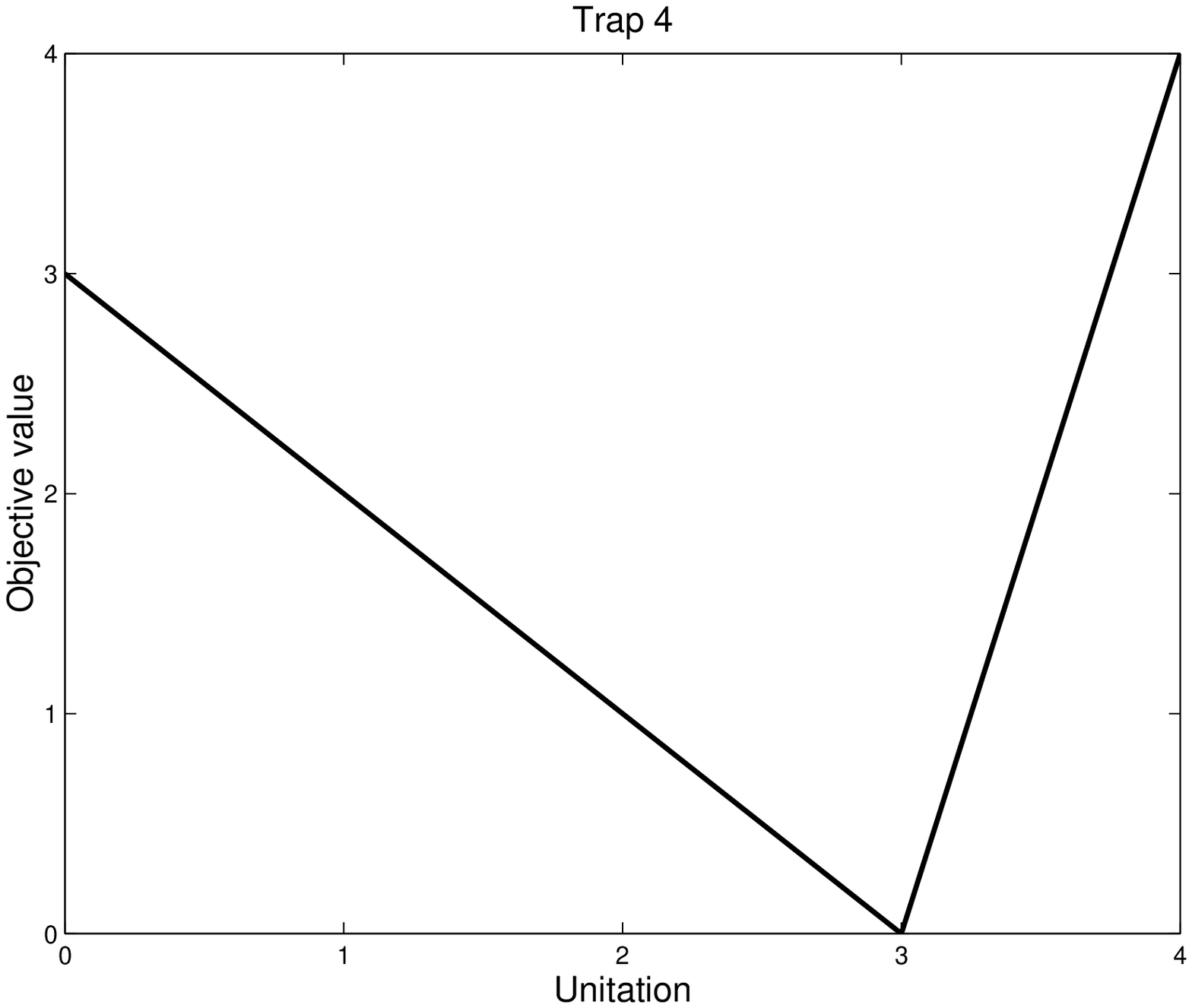, width=2.1in, height=2in}
 \epsfig{figure=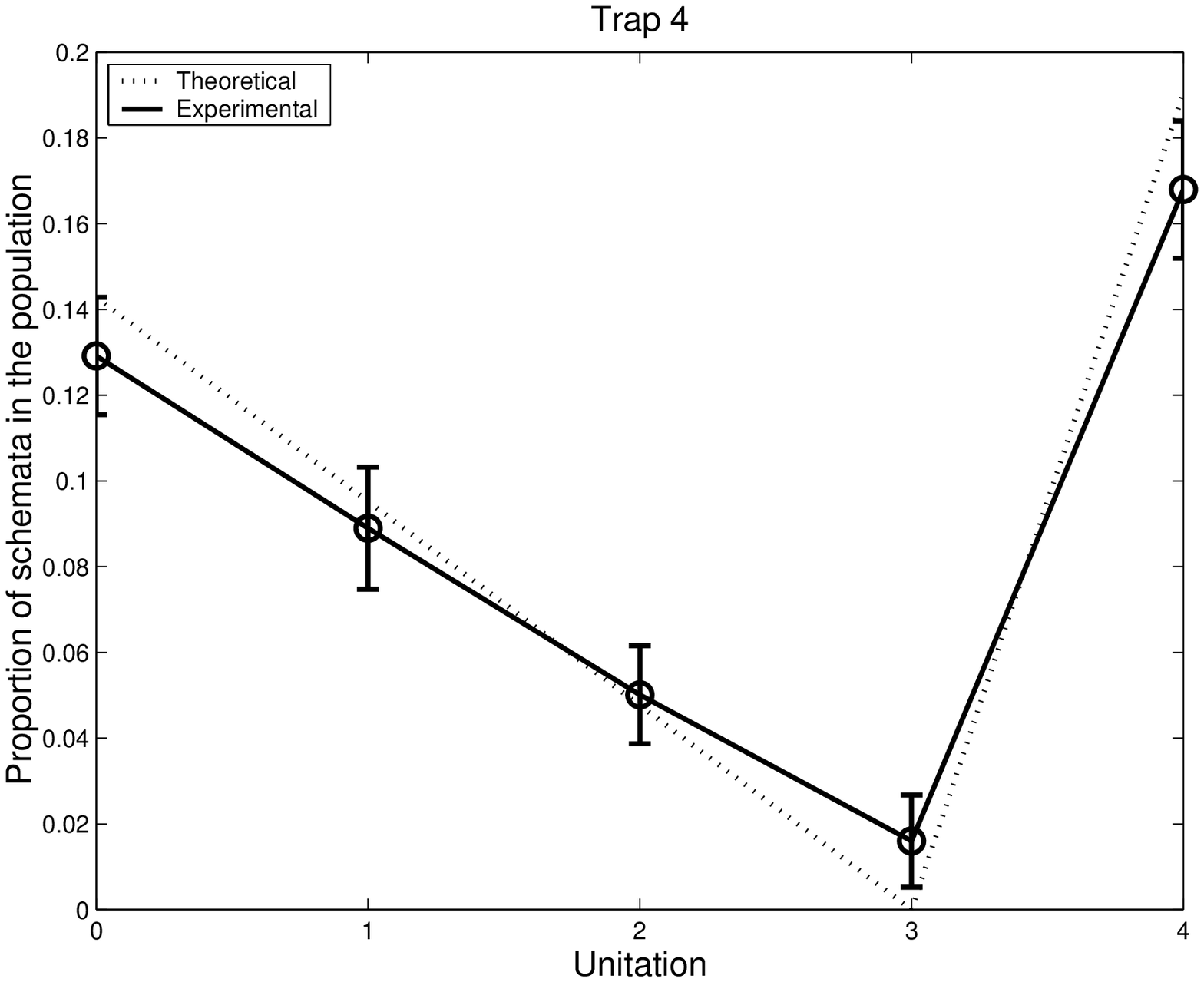, width=2.1in, height=2in}
 \caption{On left, trap--4. On right, theoretical and experimental fitness sample.}\label{ftrap4p}
\end{center}
\end{figure}

Figure~\ref{ftrap4p}--left depicts the trap--4 function we use in this experiment. The first key
question in these experiments is whether or not during the first generation of niching, the niching
method correctly samples the actual fitness function. We define a unitation as a function which
counts the number of 1's in a chromosome. Given an order $k$ trap, the theoretical proportion of a
schema $s$ with a unitation of $i$ is calculated as follows:
\begin{equation}
p(s=i) = \frac{C(k,i) \times f(s=i)}{\sum_{j=0}^k C(k,j) \times f(s=j)}
\end{equation}
Figure~\ref{ftrap4p}--right depicts the theoretical and experimental proportion of the schemas,
where it is clear that the building blocks exist in proportion to their schema fitness.

\begin{figure} [h!]
\begin{center}
 \epsfig{figure=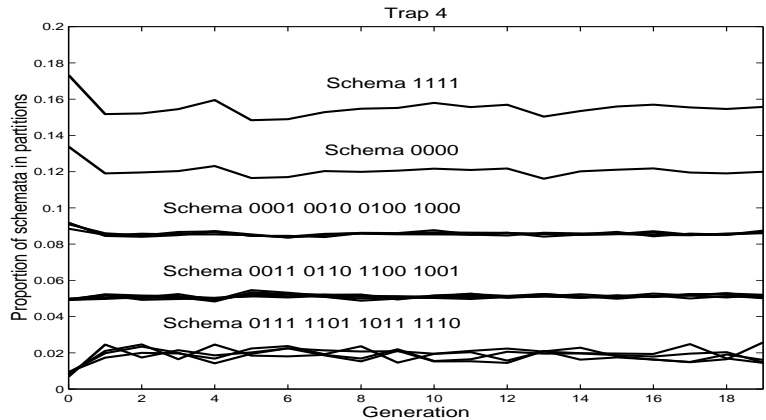, width=4in, height=2.2in}
 \caption{The modified trap function 4 in a changing environment.}\label{ftrap4s}
\end{center}
\end{figure}

Once we ensure that the building block exists in proportion to their fitness, the second question
to answer is whether the niching method we propose is sufficient to maintain the relative
proportion of the different schemas correctly. Figure~\ref{ftrap4s} confirms this behavior, where
one can see that the niching method was able to maintain the relative proportion of the different
schemas.

We can conclude from the previous experiment that the niching method is successful in maintaining
the schemas. These results do not require any additional experiments in a changing environment
where the environment switches between the already maintained schemas. For example, if the
environment is switching between schema 0000 and schema 1111 as the global optima, the previous
results is sufficient to guarantee the best performance in a changing environment. One of the main
reason for that is the environment is only manipulating the good schemas. However, what will happen
if the bad schemas become the good ones and the environment is reversing the definition of a good
and bad schemas. We already know that the maintenance of all schemas will slow down the convergence
and because of selection pressures, below average schemas will eventually disappear. Therefore, we
construct our experimental setup in a changing environment problem with two challenging problems
for niching. The first problem alters the definition of above and below average schemas, while the
second problem manipulates the boundaries of the building blocks (switching between two values of
k).

\section{Experiments}

We repeated each experiment 30 times with different seeds. All results are presented for the
average performance over the 30 runs. The population size is chosen large enough to provide enough
samples for the probabilistic model to learn the structure and is fixed to 5000 in all experiments.
Termination occurs when the algorithm reaches the maximum number of generations of 100. We assume
that the environment changes between generations and the changes in the environment are assumed to
be cyclic, where we tested two cycles of length 5 and 10 generations respectively. The crossover
probability is 1, and the tournament size is 16 in all experiments based on Harik's default values.

\subsection{Experiment 1}

\begin{figure} [h!]
\begin{center}
 \epsfig{figure=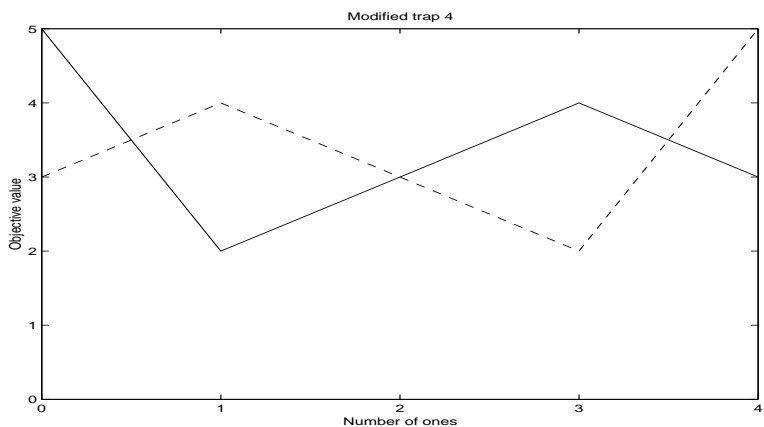, width=4in, height=2.2in}
 \caption{The modified trap function 4 in a changing environment.}\label{ftrap2}
\end{center}
\end{figure}

In the initial set of experiments, we modified the trap function of order 4 to break the symmetry
in the attractors. In this section, we choose $low=k$, $high=k+1$. Symmetry can be utilized by a
solver to easily track optima. The new function is visualized in Figure~\ref{ftrap2}. At time 0 and
in even cycles, the optimal solution is when all variables are set to 0's and the second attractor
is when the sum of 1's is equal to 3. When the environment changes during the odd cycles, the new
solution is optimal when all variables are set to 1's and the new deceptive attractor is when the
sum of 1's is 1 or alternatively the number of 0's is 3.

\begin{figure} [h!]
\begin{center}
 \epsfig{figure=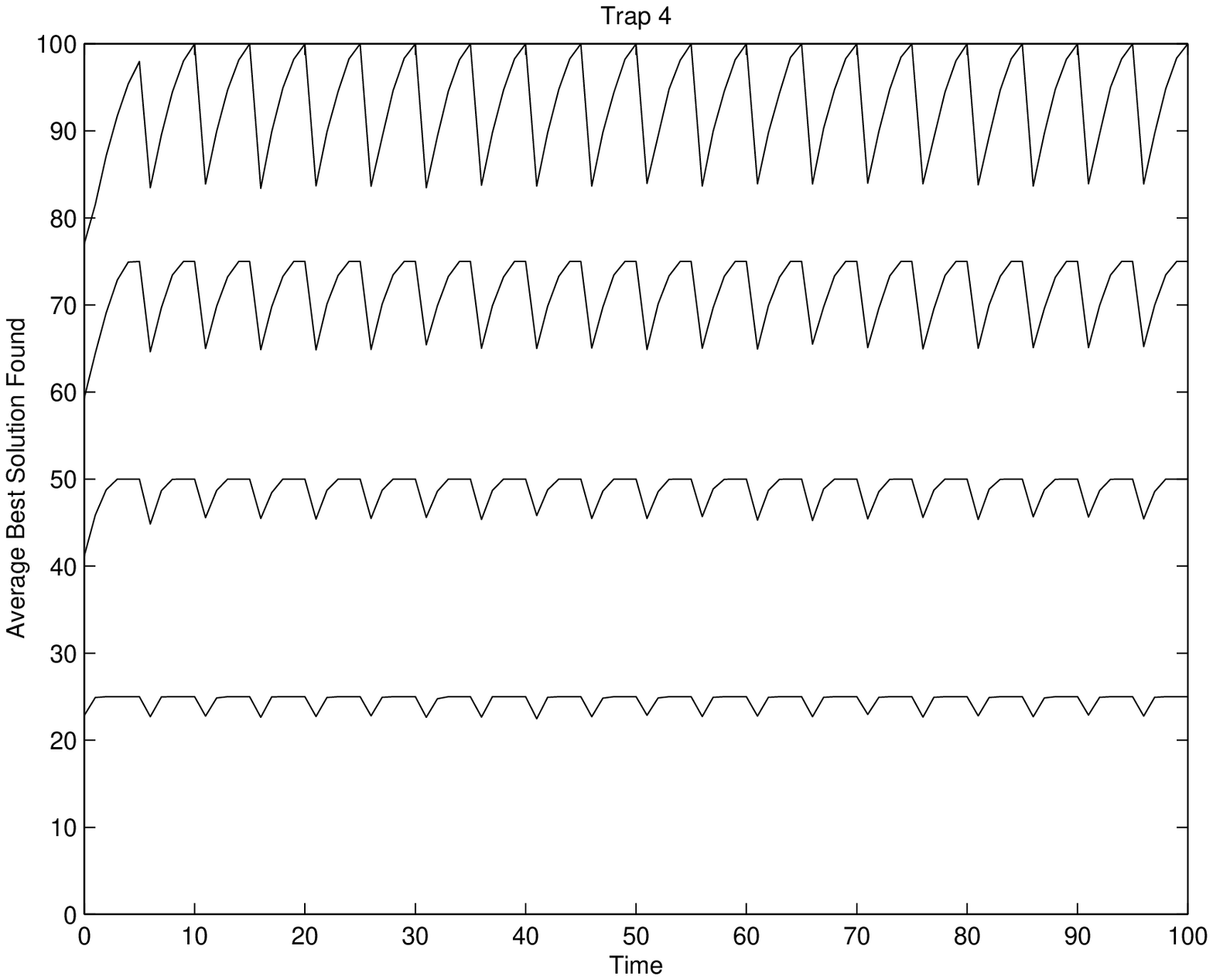,  width=2.1in, height=2in}
 \epsfig{figure=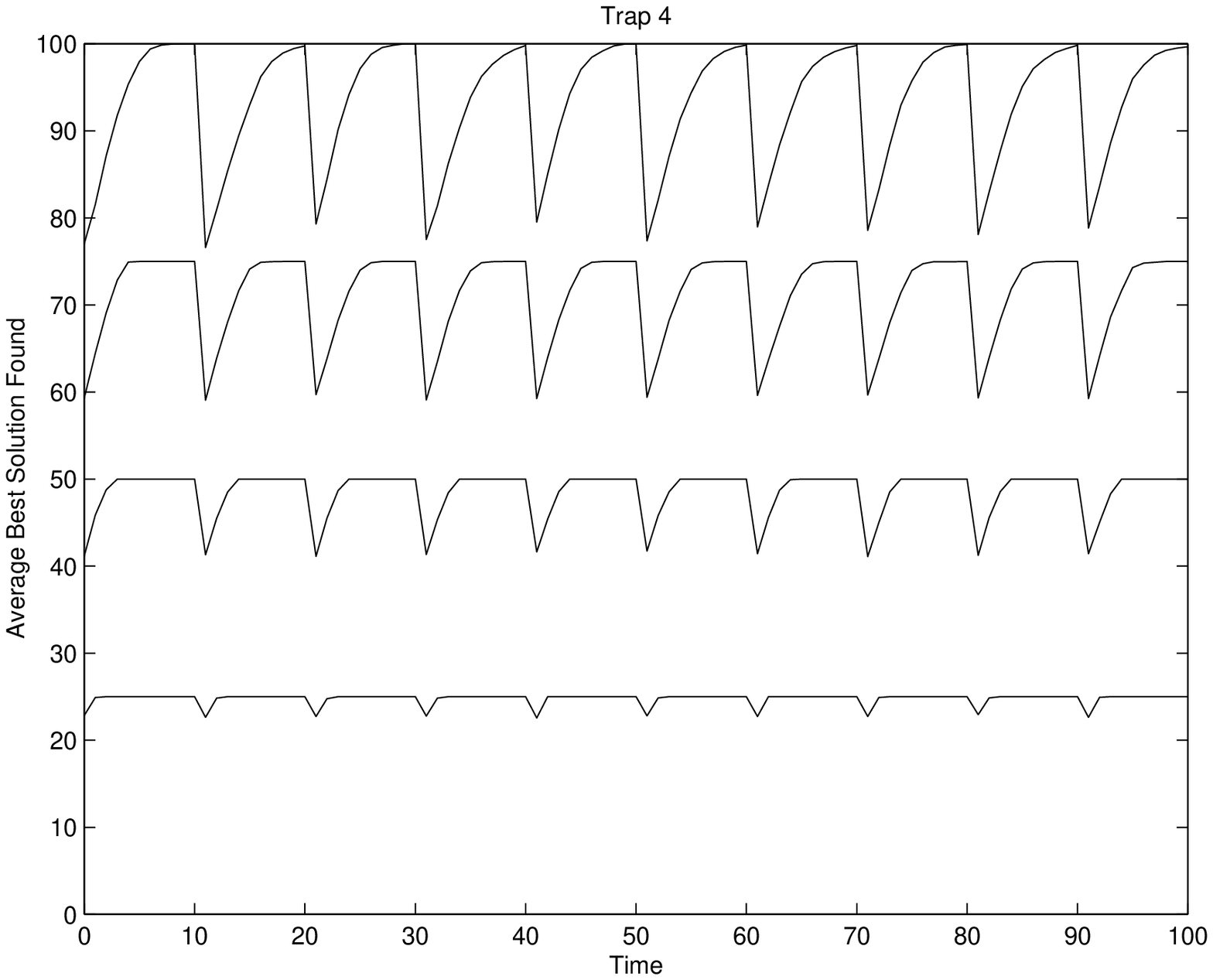,  width=2.1in, height=2in}

 \epsfig{figure=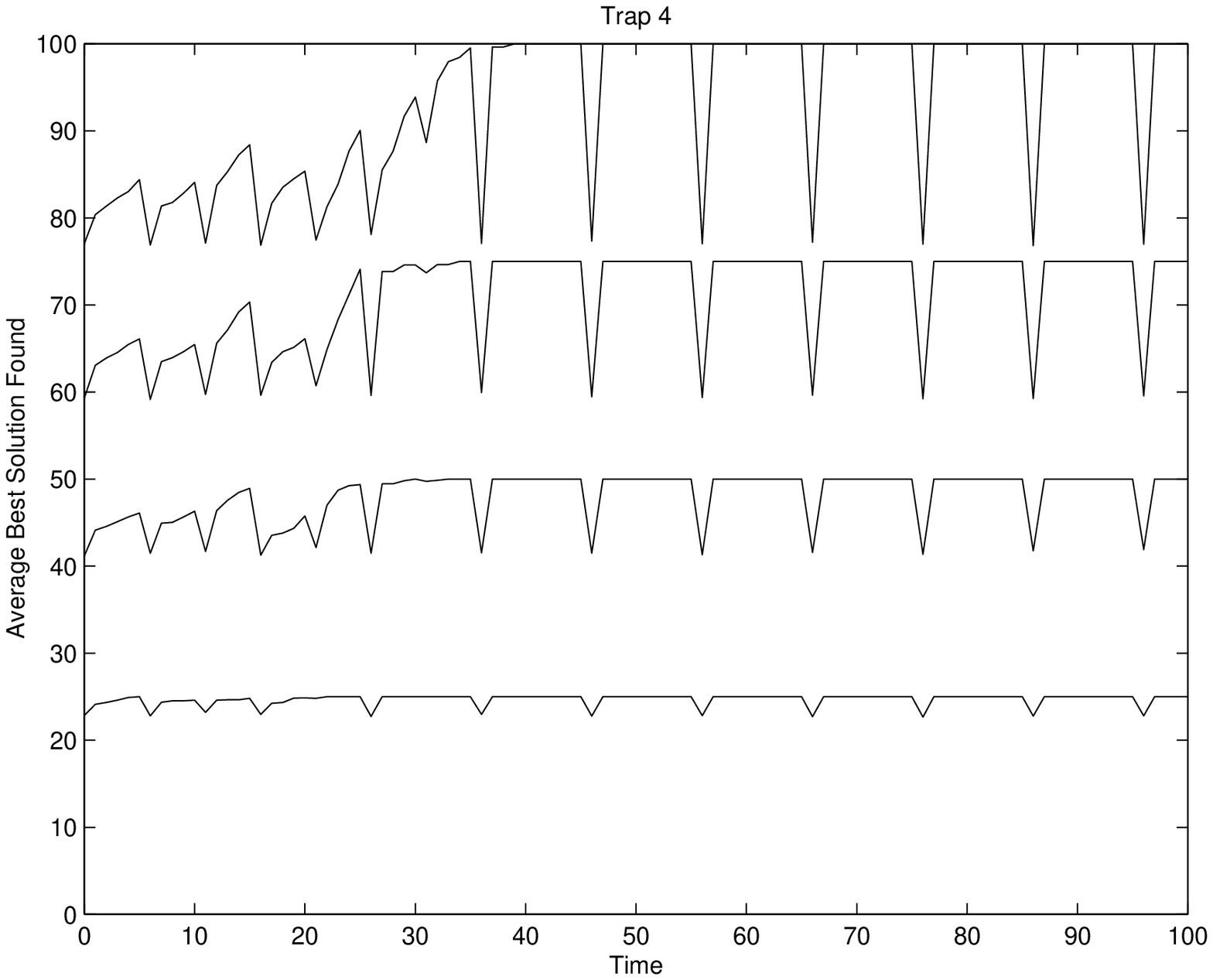,  width=2.1in, height=2in}
 \epsfig{figure=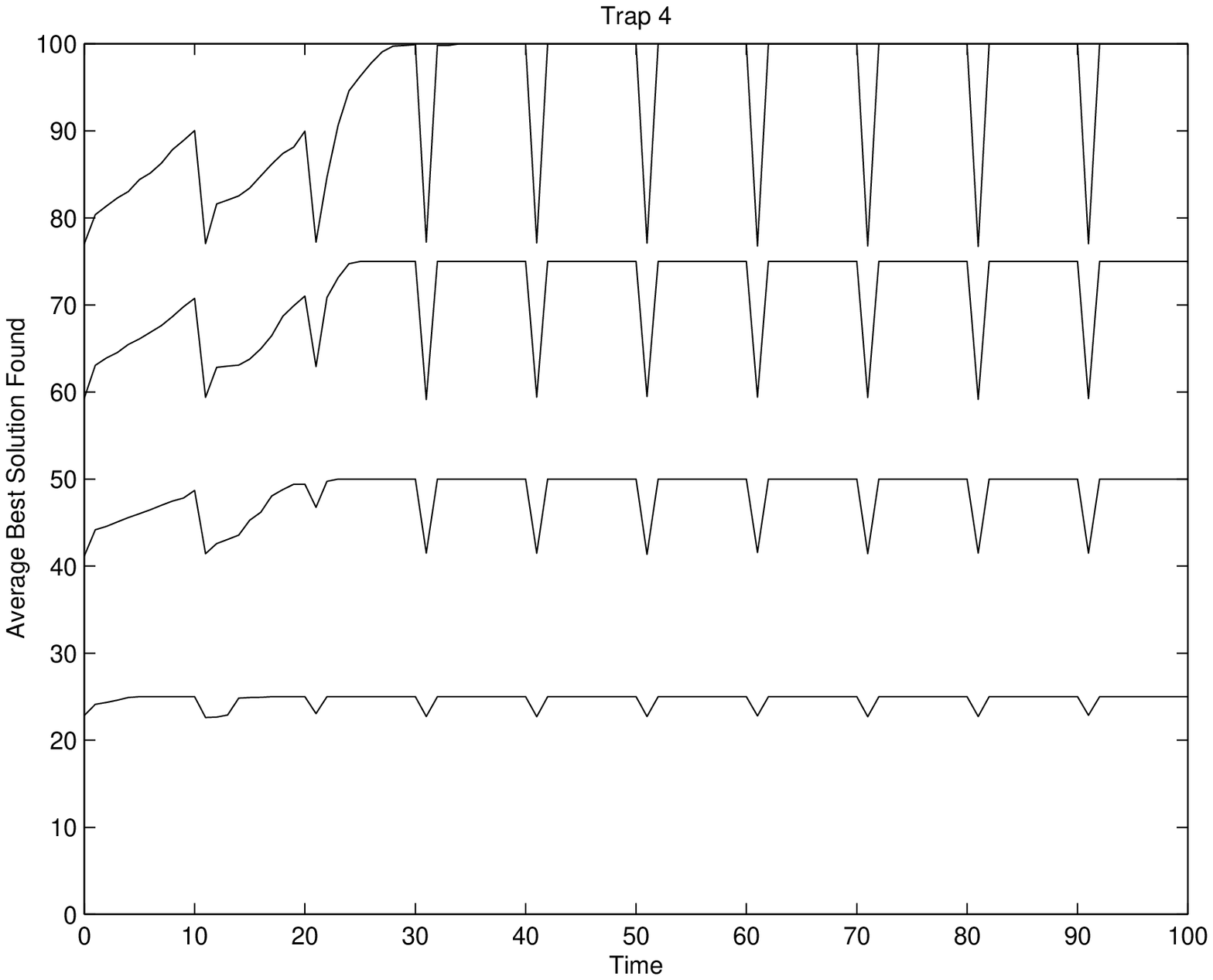,  width=2.1in, height=2in}

 \epsfig{figure=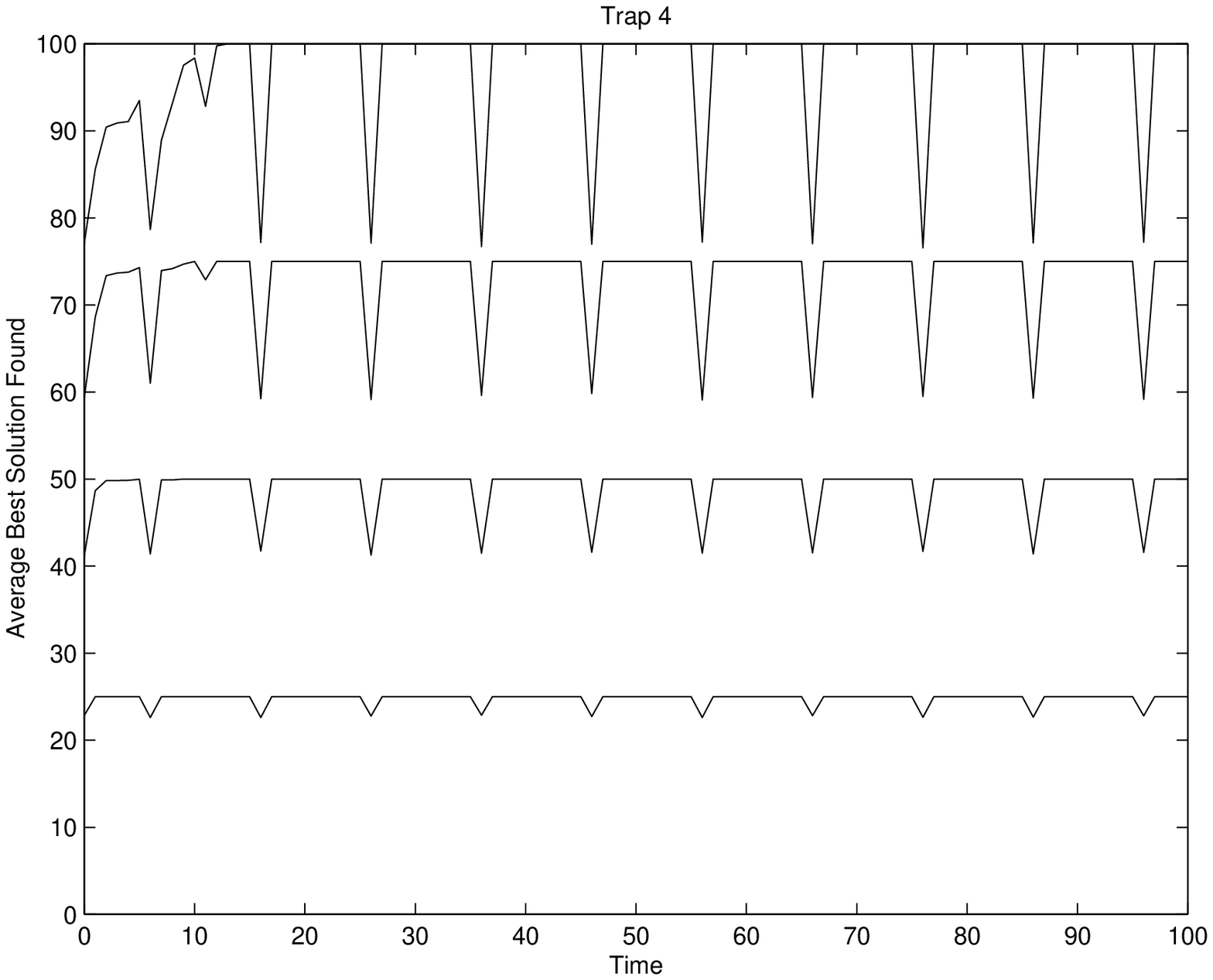,  width=2.1in, height=2in}
 \epsfig{figure=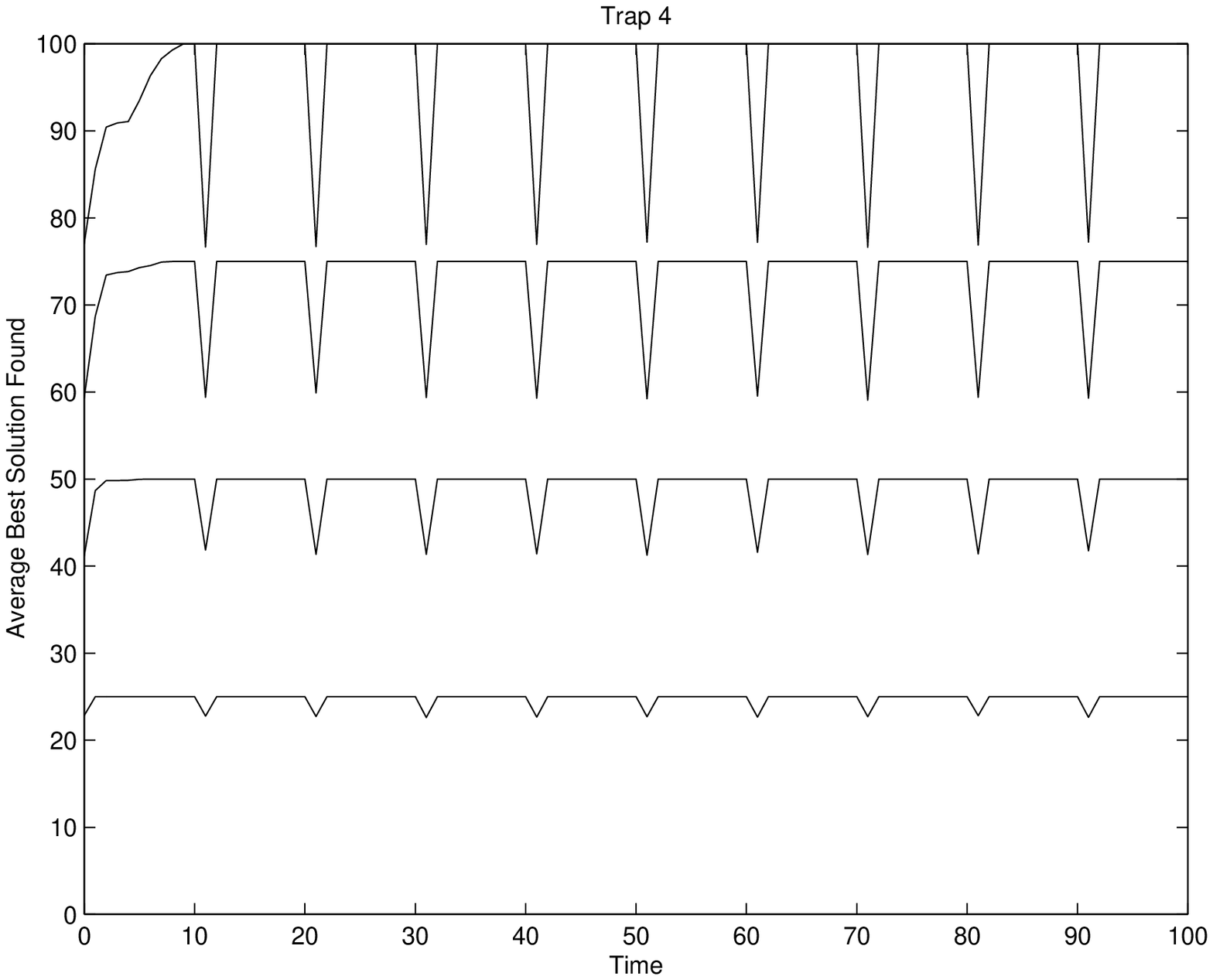,  width=2.1in, height=2in}

 \caption{Modified Trap 4 (left) Cycle 5 (Right)
 Cycle 10, (top) dcGA, (middle) Schem1, (Bottom) Schem2. In each graph, the four curves
correspond to 5, 10, 15, and 20 building blocks ordered from bottom up.}\label{res4}
\end{center}
\end{figure}

Figure~\ref{res4} depicts the behavior of the three methods using the modified trap--4 function. By
looking at the results for dcGA, the response rate ({\it i.e.} the time between a change and
reaching the new optimal solution) is almost at the edge of 5 for 20 building blocks. This means
that the algorithm requires on average 5 generations to get close to the new optimal. By looking at
the cycle of length 10, it becomes clearer that the algorithm takes a bit more than 5 generations
(between 6-7 generations) to converge.

When looking at Schem1, we can see that the algorithm takes longer in the first phase to get to the
optimal solution. On the average, it takes 30 generations to do so. We will call this period the
``warming up'' phase of the model. The niching method delays the convergence during this stage.
However, once the warming up stage is completed, the response rate is spontaneous; once the change
occurs, a drop occurs then the method recovers instantly and gets back to the original optima. By
comparing Schem1 and Schem2, we find the two methods are very similar except for the warming--up
stage, where Schem2, which uses the above average schemas only, has a shorter time to warm-up than
Schem1.

\subsection{Experiment 2}

\begin{figure} [h!]
\begin{center}
 \epsfig{figure=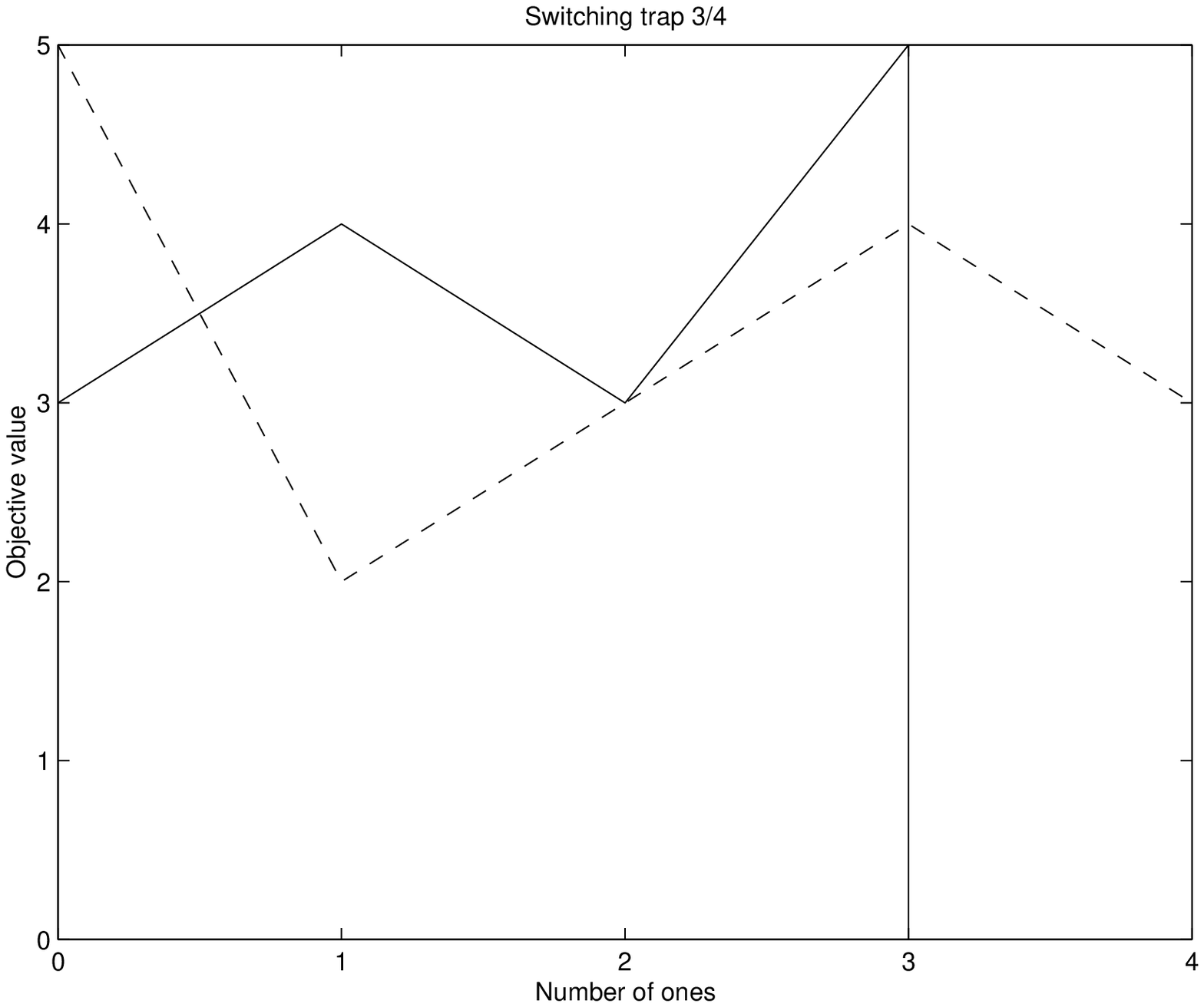, width=4in, height=2.2in}
 \caption{The switching trap function with k=3,4 in a changing environment.}\label{ftrap3}
\end{center}
\end{figure}

In this experiment, we subjected the environment under a severe change from linkage point of view.
Here, the linkage boundary changes as well as the attractors. As being depicted in
Figure~\ref{ftrap3}, the environment is switching between trap--3 with all optima at 1's and
trap--4 with all optima at 0's. Moreover, in trap--3, a deceptive attractor exists when the number
of 1's is 1 while in trap--4, a deceptive attractor exists when the number of 1's is 3. This setup
is tricky in the sense that, if a hill climber gets trapped at the deceptive attractor for trap--4,
the behavior will be good for trap--3. However, this hill--climber won't escape this attractor when
the environment switches back to trap-4 since the solution will be surrounded with solutions of
lower qualities. This setup tests also whether any of the methods is behaving similar to a
hill--climber.

\begin{figure} [h!]
\begin{center}
 \epsfig{figure=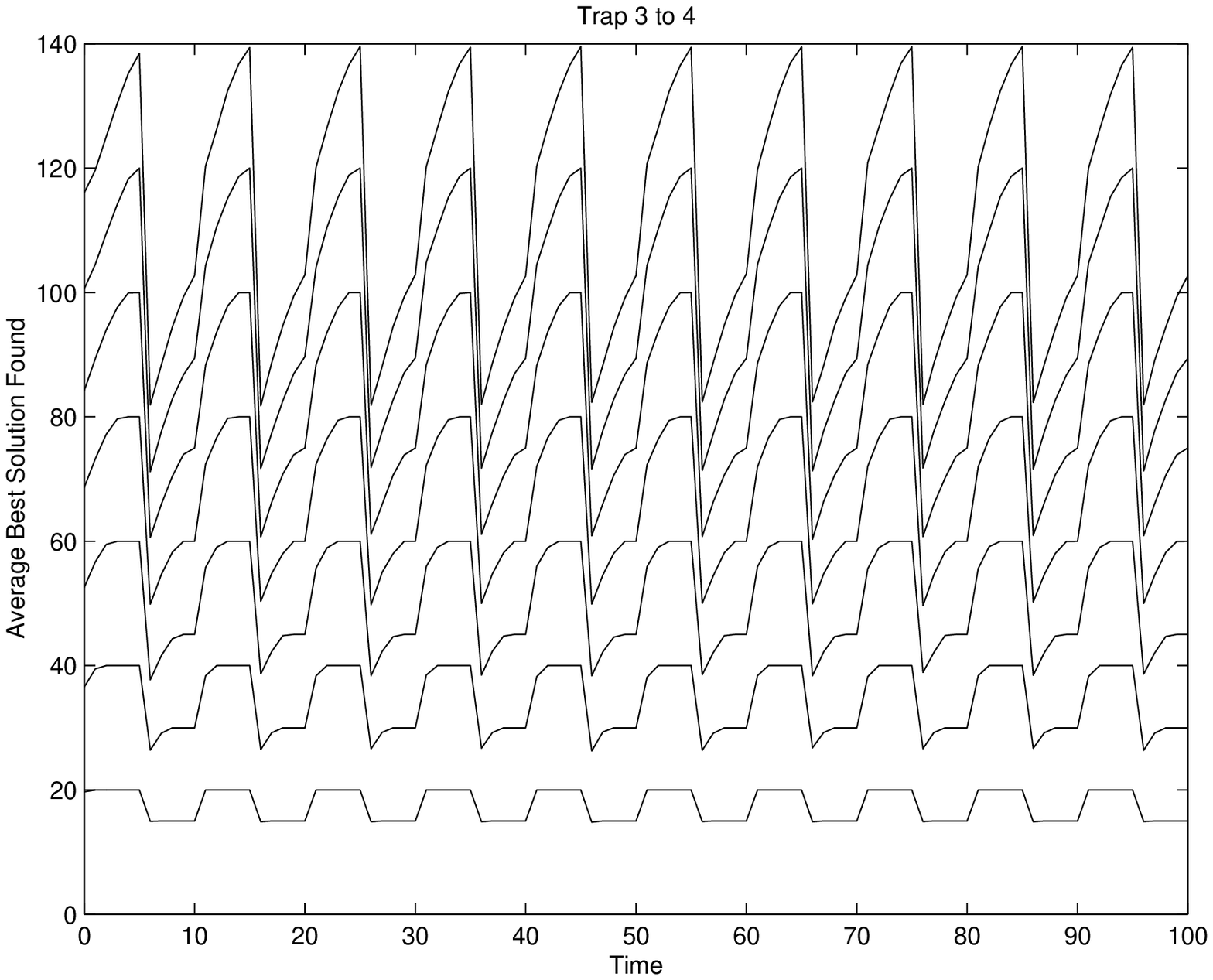,  width=2.1in, height=2in}
 \epsfig{figure=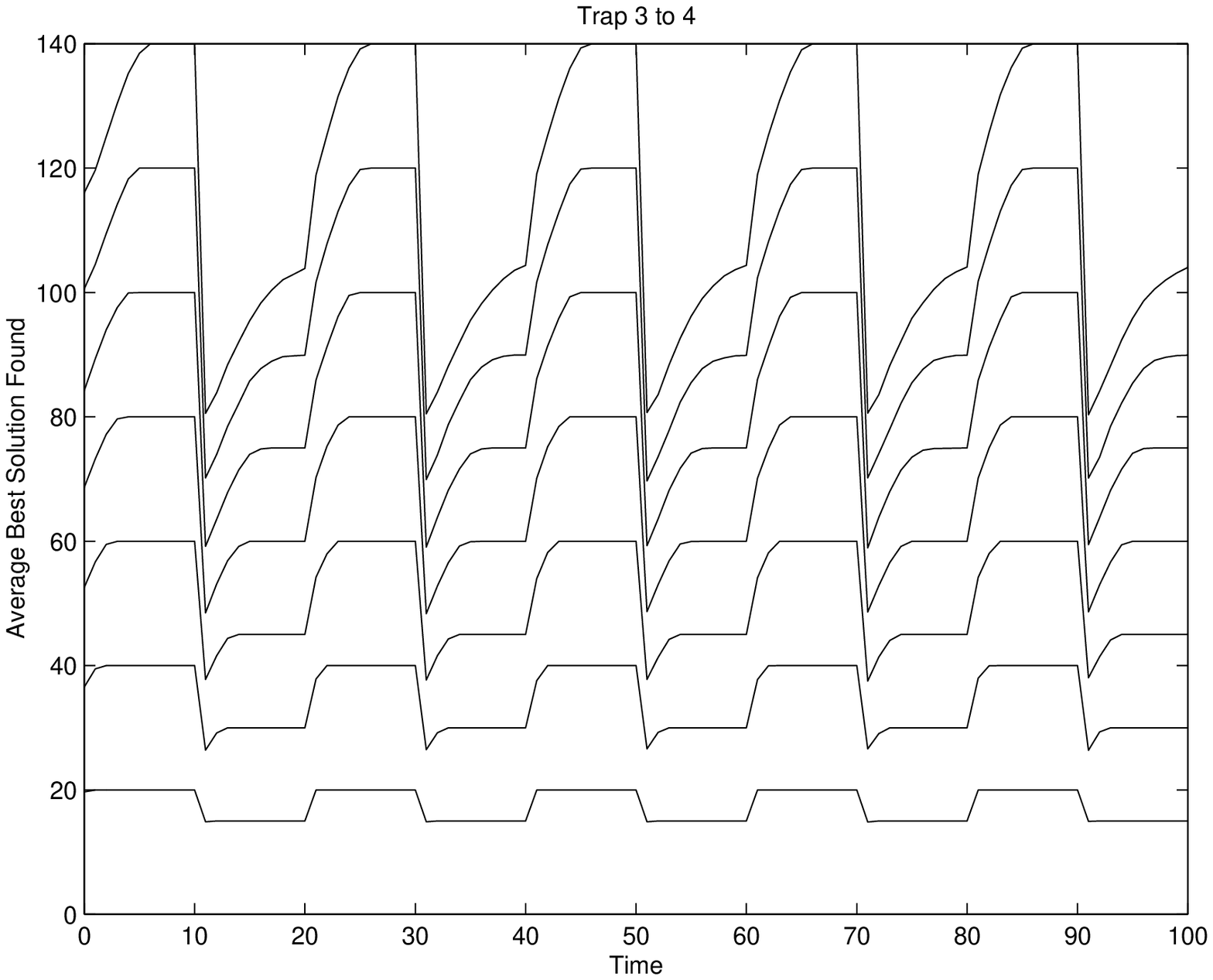,  width=2.1in, height=2in}

 \epsfig{figure=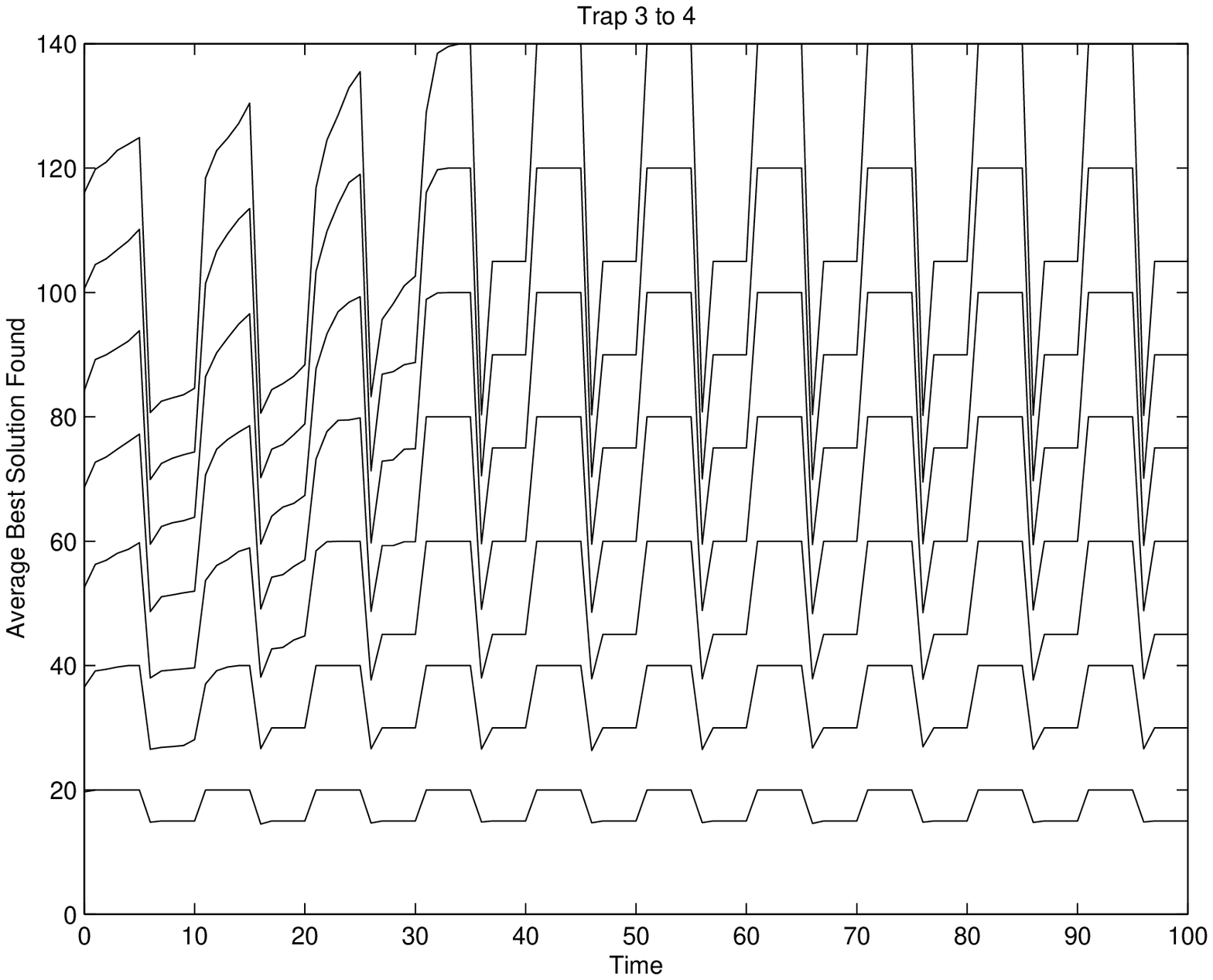,  width=2.1in, height=2in}
 \epsfig{figure=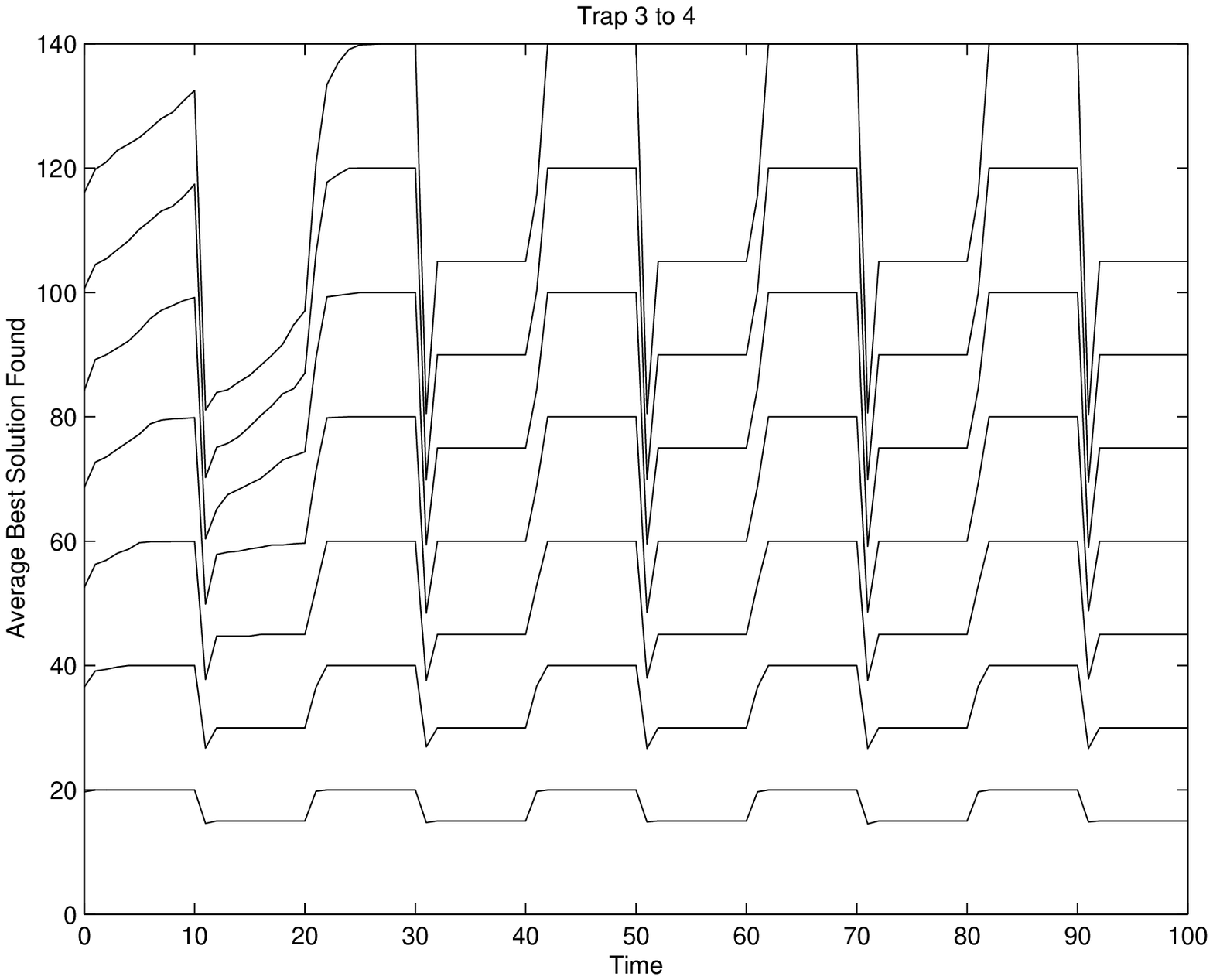,  width=2.1in, height=2in}

 \epsfig{figure=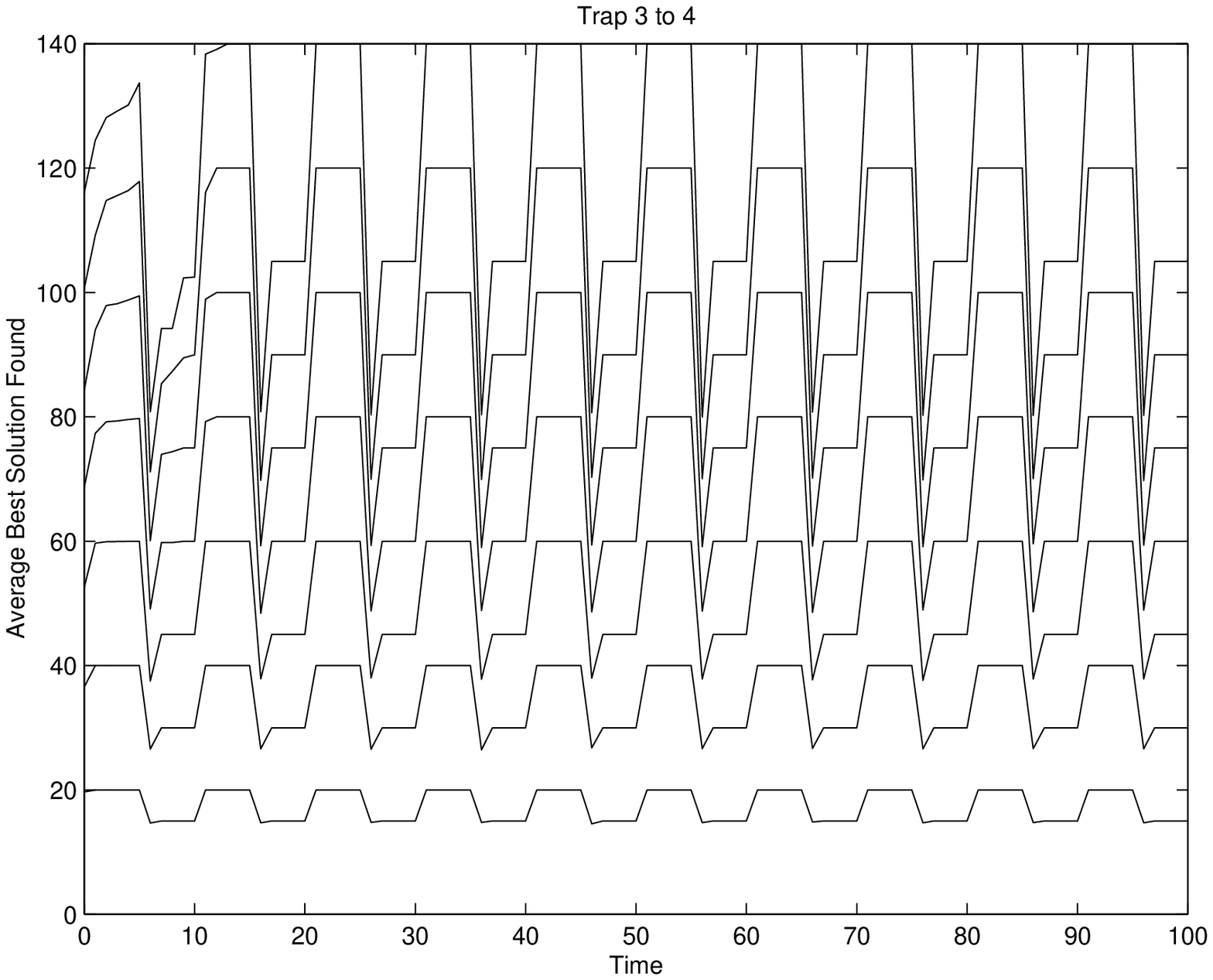,  width=2.1in, height=2in}
 \epsfig{figure=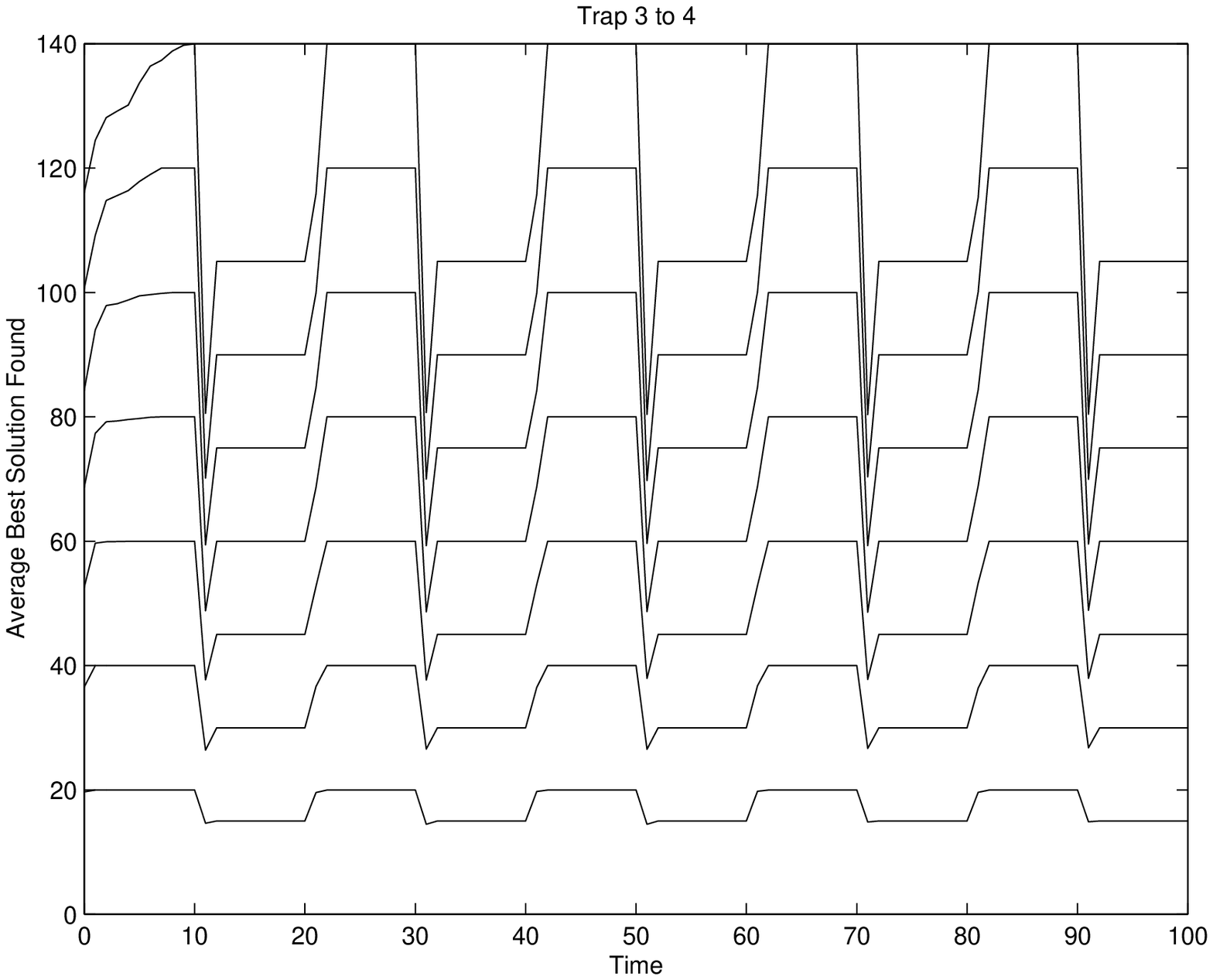,  width=2.1in, height=2in}
 \caption{Switching Trap 3--4 (left) Cycle 5 (Right)
 Cycle 10, (top) dcGA, (middle) Schem1, (Bottom) Schem2. In each graph,
 the seven curves correspond to strings of length 12, 24, 36, 48, 60, 72,
 and 84 bits ordered from bottom up.}\label{res5}
\end{center}
\end{figure}

Figure~\ref{res5} shows the performance of dcGA, Schem1, and Schem2. We varied the string length
between 12 and 84 in a step of 12 so that the string length is dividable by 3 and 4 (the order of
the trap). For example, if the string length is 84 bits, the optimal solution for trap 4 is $5
\times \frac{84}{4} = 105$ and for trap 3 is $5 \times \frac{84}{3} = 120$. Therefore, the
objective value will alternate between these two values at the optimal between cycles. The results
in Figure~\ref{res5} are very similar to the previous experiment. The dcGA method responds
effectively to environmental changes but Schem1 and Schem2 respond faster. Also, the warming up
period for Schem1 is longer than Schem2.

\section{Conclusion}

In this paper, we presented a niching method based on an automatic problem decomposition approach
using competent GAs. We have demonstrated the innovative idea that niching is possible on the
sub--structural level despite that the learning of these sub--structures is adaptive and may be
noise. We tested changes where the functions maintain their linkage boundaries but switches between
optima, as well as drastic changes where the functions change their optima simultaneously with a
change in their linkage boundaries. In all cases, niching on the sub--structural level is a robust
mechanism for changing environments.

\section{Acknowledgment}

{\small This work was sponsored by the Air Force Office of Scientific Research, Air Force Materiel
Command, USAF, under grant F49620-00-0163 and F49620-03-1-0129, by the Technology Research,
Education, and Commercialization Center (TRECC), at UIUC by NCSA and funded by the Office of Naval
Research (grant N00014-01-1-0175), the National Science Foundation under ITR grant DMR-99-76550 (at
Materials Computation Center), ITR grant DMR-0121695 (at CPSD), the Dept. of Energy under grant
DEFG02-91ER45439 (at Fredrick Seitz MRL), the University of New South Wales SSP Scheme, and the
Australian Research Council (ARC) Centre on Complex Systems grant number CEO0348249.}

\bibliographystyle{plain}

\end{document}